\documentclass{article}

\usepackage{PRIMEarxiv}

\usepackage[utf8]{inputenc} 
\usepackage[T1]{fontenc}    
\usepackage{hyperref}       
\usepackage{url}            
\usepackage{booktabs}       
\usepackage{amsfonts}       
\usepackage{nicefrac}       
\usepackage{microtype}      
\usepackage{lipsum}
\usepackage{fancyhdr}       
\usepackage{graphicx}       
\usepackage{bm}             
\usepackage{amsmath}
\usepackage{subcaption}
\usepackage{multirow}
\usepackage{array}
\usepackage{wrapfig}

\title{Toward Understanding the Disagreement Problem in Neural Network Feature Attribution
}

\author{
  Niklas Koenen$^{1,2}$ and Marvin N. Wright $^{1,2,3}$
\and
$^1$Leibniz Institute for Prevention Research \& Epidemiology – BIPS  
\and
$^2$Faculty of Mathematics and Computer Science, University of Bremen
\and
$^3$Department of Public Health, University of Copenhagen\\
\texttt{koenen@leibniz-bips.de} \\
}

\begin{document}
\maketitle

\begin{abstract}
In recent years, neural networks have demonstrated their remarkable ability to discern intricate patterns and relationships from raw data. However, understanding the inner workings of these black box models remains challenging, yet crucial for high-stake decisions. Among the prominent approaches for explaining these black boxes are feature attribution methods, which assign relevance or contribution scores to each input variable for a model prediction. Despite the plethora of proposed techniques, ranging from gradient-based to backpropagation-based methods, a significant debate persists about which method to use. Various evaluation metrics have been proposed to assess the trustworthiness or robustness of their results. However, current research highlights disagreement among state-of-the-art methods in their explanations. Our work addresses this confusion by investigating the explanations' fundamental and distributional behavior. Additionally, through a comprehensive simulation study, we illustrate the impact of common scaling and encoding techniques on the explanation quality, assess their efficacy across different effect sizes, and demonstrate the origin of inconsistency in rank-based evaluation metrics.
\end{abstract}



\section{Introduction}

Over the past decade, one specific class of machine learning models has been rapidly integrated into our daily lives: neural networks. Thanks to increasing computational power and resources, it has become possible to control these exceptionally flexible and highly parameter-rich models. Their remarkable success spans from image recognition to financial forecasts and disease detection \cite{lecun_deep_2015,bengio2021deep}. Nevertheless, this black box and its prediction-making process are challenging or even impossible for humans to fully understand. 

Building upon this question of explaining a model's prediction, the term \emph{explainable artificial intelligence} (XAI) has emerged, and a fast-growing research area has been established. Many methods have been proposed to explain black box models, ranging from intrinsic and self-explaining neural networks \cite{alvarez_melis_towards_2018,tabnet_2021} to concept-recognizing \cite{bau_network_2017,kim_interpretability_2018} and per\-tur\-ba\-tion-based approaches \cite{IVANOVS2021228,petsiuk_rise_2018}. However, the most well-known and commonly used group for post-hoc explanations consists of \emph{feature attribution methods}. For a trained model, they assign relevance or contribution scores to the input variables, thus indicating the features or components on which the model bases its prediction. They have become known primarily for their type of visualization as heatmaps or saliency maps for image data. Nevertheless, most methods can also be applied to other data types, such as tabular data, due to the feature-individual attributions.

Driven by an increasingly diverse collection of methods \cite{simonyan_deep_2013,shrikumar_not_2016, smilkov_smoothgrad_2017, srinivas_full-gradient_2019,sundararajan_axiomatic_2017,bach_pixel-wise_2015,shrikumar_learning_2017}, feature attribution research has recently shifted away from method development and toward the question of which method is the 'best'. However, this question can be answered differently depending on which aspect is being asked. In this sense, numerous partly heuristic evaluation metrics have been proposed measuring, e.g., an explanation's trustworthiness/fidelity \cite{alvarez_melis_towards_2018,samek_evaluating_2017,ancona_towards_2018,yeh_fidelity_2019,dasgupta_framework_2022}, robustness \cite{krishna_disagreement_2022,yeh_fidelity_2019}, complexity \cite{nguyen_quantitative_2020,bhatt_evaluating_2020}, or monotonicity \cite{arya_one_2019}. Additionally, feature attribution methods have been compared in benchmarks using these metrics, mostly concluding that the choice is either model or dataset-dependent \cite{adebayo_sanity_2018,liu_synthetic_2021,agarwal_openxai_2022,yang_benchmarking_2019}. Bhatt et al. \cite{bhatt_evaluating_2020} even propose a meta-metric based on aggregating various metrics. 

\begin{figure}[!t]
    \centering
    \includegraphics[width = \textwidth]{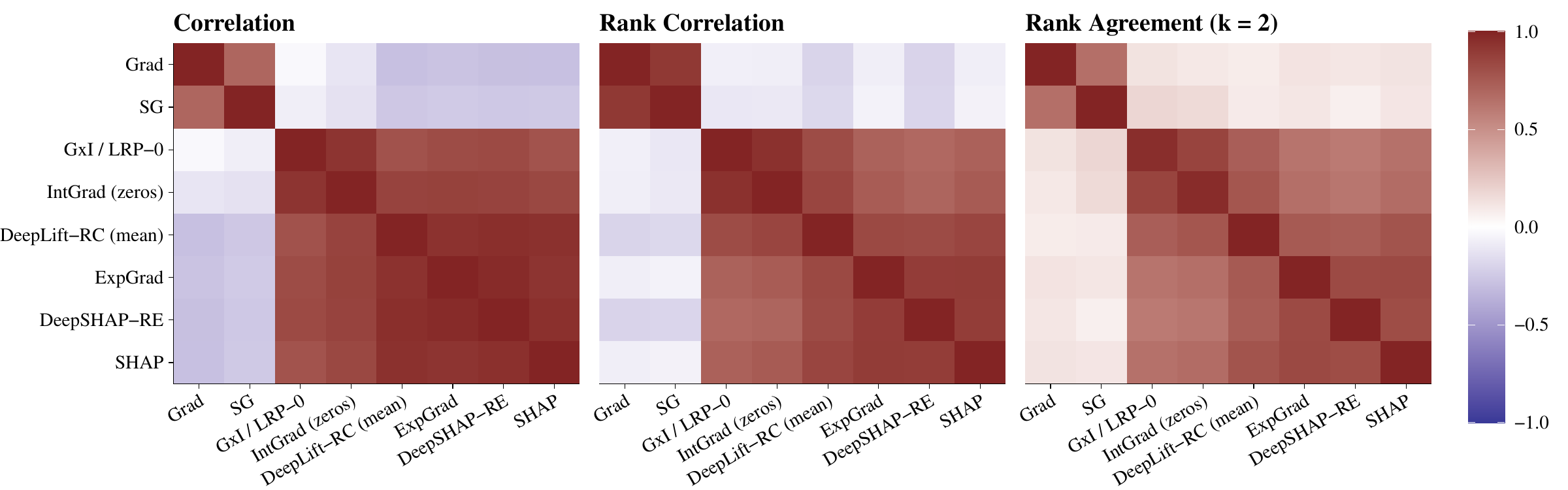}
    \caption{Pairwise comparison of state-of-the-art feature attribution methods (see Sec.~\ref{Sec:background_methods} for the method descriptions) on the COMPAS dataset using (left) the mean feature-correlation, (middle) mean instance-wise Kendall rank correlation, and mean rank agreement of the two most important features (details can be found in Appendix~\ref{App:COMPAS}).}
    \label{fig:intro_agreement}
\end{figure}

Most of these evaluation metrics are based on a (simulated) information elimination of highly attributed features and the resulting observable change in the model's prediction \cite{samek_evaluating_2017,ancona_towards_2018}, and do not consider the explanation's distributional behavior. Apart from prediction-grounded evaluations, the magnitude of relevances is also used for pairwise comparisons of the method's local explanations. In this context, Krishna et al. \cite{krishna_disagreement_2022} postulated the so-called \emph{disagreement problem}, as many state-of-the-art methods differ significantly in their assignment of important features. The authors defined rank-based metrics on the basis of Neely et al. \cite{neely2021order}, e.g., the rank agreement or rank correlation, and showed the disagreement using real-world datasets. We reproduce their comparison on the COMPAS dataset \cite{bao2021its} and also observe this disagreement (see Figure.~\ref{fig:intro_agreement} middle and right). However, we claim this disagreement is not mainly caused by different explanation qualities but rather by how the effects are measured for many state-of-the-art methods. Moreover, we demonstrate that the magnitude of the relevance is strongly affected by an implicitly or explicitly chosen baseline value causing the method to answer different questions. For example, we can see the high correlation of the feature-wise distributions for non-plain gradient-based methods in the left heatmap in Figure.~\ref{fig:intro_agreement}.

Our work aims to fundamentally understand the most prominent methods for feature attribution in neural networks while uncovering their limitations and resulting misinterpretations caused by visualizations and the choice of the baseline value. For example, a relevant feature can become less or even irrelevant for another baseline. To achieve this, we categorize these methods into four groups based on their underlying explanation target and analyze their distributional behavior. We demonstrate situations where certain methods fail to deliver adequate attributions through simulation studies employing a known data generation process for tabular data. Furthermore, we investigate the impacts of various common data preprocessing techniques for continuous and categorical features on the explanation methods. Using the simulations, we clarify whether a feature attribution method can correctly attribute relevance to the prediction on a local level and whether rank-based aggregated relevance can be used as a global feature importance measure.


\section{Background and Related Work}


\subsection{Preliminaries}

To contextualize the topic of our work within the multitude of interpretation methods for machine learning models, we rely on the taxonomic classifications proposed by Doshi-Valez and Kim \cite{doshi-velez_towards_2017} and, specifically for neural networks, the framework provided by Zhang et al. \cite{zhang_survey_2021}. Accordingly, feature attribution methods are considered (semi-)local post-hoc techniques, as they require a trained model (i.e., all parameters and internal structures remain unchanged) and explain the prediction of a single instance, e.g., an image or a patient. The term \emph{feature attribution} originates from the fact that these methods assign relevance or contribution scores on a prediction to each feature. This means that for a model $f$ and an instance $\bm{x} \in \mathbb{R}^p$, a feature attribution method results in a vector $\bm{r} = (r_1, \ldots, r_p)^T \in \mathbb{R}^p$ of feature-wise relevances. Ideally, they should provide a decomposition of the prediction (or a proportional objective) into individual feature-wise effects, also referred to in the literature as local accuracy \cite{lundberg_unified_2017}, completeness \cite{montavon_explaining_2017}, or summation-to-delta property \cite{shrikumar_learning_2017}, i.e., with $r_0 \in \mathbb{R}$
\begin{align}\label{background:local_accuracy}
    f(\bm{x}) = r_0  + \sum_{i = 1}^p r_i.
\end{align}
Even if not all methods fulfill this property exactly, a reasonable feature attribution method should comply with this fundamental principle or strive to approximate it. In the case of a linear model -- which, in principle, is a neural network with only one dense layer and linear activation -- the relevances for an ideal method should be proportional to the product of the regression coefficients and the feature values, i.e., $\beta_i\, x_i$. In this notation, Shapley values provide proportional explanations, as they estimate the local effect against the constant marginal effect $\beta_i\, x_i - \beta_i \mathbb{E}[X_i]$.


\subsection{Methods}
\label{Sec:background_methods}

This paper focuses on feature attribution methods specifically designed for neural networks, i.e., model-specific methods, which allow an application to image and tabular data. The pioneering work in this field is the gradient method (Grad) introduced by Simonyan et al. \cite{simonyan_deep_2013}, which computes the feature-wise partial derivatives from the output to the respective input features. Originally applied to image data, the sum or maximum of absolute values across the color channels are calculated and became famous as saliency maps (Saliency). Subsequently, further methods emerged specifically for convolutional neural networks and ReLU networks \cite{springenberg_striving_2014,selvaraju_grad-cam_2017,zeiler_visualizing_2014}. These variations primarily differ in their computation of gradients within activations or their incorporation of the gradients to bias terms \cite{srinivas_full-gradient_2019}. However, Smilkov et al. \cite{smilkov_smoothgrad_2017} critique the visually noisy nature of saliency maps, leading to the development of SmoothGrad (SG). This approach estimates the average gradient of Gaussian-disturbed inputs, resulting in a sharper appearance of the saliency map. Alternatively, Adebyo et al. \cite{adebayo_local_2018} proposed computing the variance instead of the average. Notably, these methods predominantly rely on the plain gradients, which, from a mathematical perspective, do not provide a direct decomposition but rather highlight the features' output sensitivity.

The first approach toward approximating the decomposition of the prediction $f(\bm{x})$ was introduced through the backpropagation-based method layer-wise relevance propagation (LRP) by Bach et al. \cite{bach_pixel-wise_2015}. LRP starts its process from the output prediction, systematically redistributing relevances layer by layer to the lower layers using predefined rules until reaching the input layer. Shrikumar et al. \cite{shrikumar_learning_2017} further advanced this concept with their deep learning important features (DeepLIFT) method, incorporating a reference value $\bm{\tilde{x}}$ (also called the baseline value) to achieve a decomposition of $f(\bm{x}) - f(\bm{\tilde{x}})$. Integrated gradient (IntGrad) \cite{sundararajan_axiomatic_2017}, sharing the same objective of decomposition as DeepLIFT, integrates the gradients along a path from $\bm{x}$ to the reference value $\bm{\tilde{x}}$. More recent techniques such as DeepSHAP and expected gradient (ExpGrad) \cite{lundberg_unified_2017,chen2021explaining} -- also known as GradSHAP -- employ multiple reference values for an explanation, bridging to Shapley values \cite{shapley1953value} by aiming for the decomposition the prediction regarding the expected prediction $f(\bm{x}) - \mathbb{E}[f(X)]$.


\subsection{Evaluation Metrics}

Evaluation metrics for feature attribution methods in current XAI research are subject to a broad debate. Doshi-Velez and Kim \cite{doshi-velez_towards_2017} categorize these metrics into human-grounded and function-grounded. The former group assesses the explanation quality based on human judgments, measuring the overall comprehensibility of non-experts. On the other hand, function-grounded metrics consist of mathematically defined criteria that can be measured without human interactions. Within this group of evaluation metrics, a prominent subgroup focuses on verifying faithfulness \cite{alvarez_melis_towards_2018,yeh_fidelity_2019}. These metrics measure the extent to which highly relevant features based on the explanation method also crucially influence the model's predictive power. They usually measure the loss change \cite{montavon_explaining_2017} or other correlations between the prediction drop and the explanations \cite{ancona_towards_2018,yeh_fidelity_2019} when the most or least relevant features are removed. Usually, the most or least relevant features are determined based on the explanations' magnitude ignoring the sign, which is consistent with the rankings of the absolute values. In this context, 'removing a feature' mostly means simulating its absence, e.g., by setting it to zero or another baseline, conditional sampling \cite{Fong_2017_ICCV}, or retraining the entire model without this feature \cite{hooker_2019}. Although all these evaluation metrics justify the explanation method's ability to detect highly decisive features, researchers have found that they are inconsistent and seem to measure different aspects \cite{tomsett_sanity_2020,gevaert_evaluating_2022}. This inconsistency has recently become known as the disagreement problem \cite{krishna_disagreement_2022,neely2021order}. 

In addition to axiomatic approaches, which verify whether methods adhere to properties \cite{khakzar_explanations_2022,sundararajan_axiomatic_2017}, e.g., local accuracy from Equation~\ref{background:local_accuracy}, our work deals with ground-truth evaluations. These evaluation methods assess the explanation's quality on synthetic datasets or injected ground-truth elements, i.e., semi-natural datasets. However, current literature primarily focuses on so-called pointing games in image data \cite{zhang_top-down_2018}, such as synthetic datasets like CLEVER-XAI \cite{arras_clevr-xai_2022} or overlaid images of prediction-relevant and irrelevant parts \cite{yang_benchmarking_2019,kim_sanity_2022,zhou_feature_2022,tjoa_quantifying_2023}.

There are comparatively fewer ground-truth analyses of model-specific feature attribution methods for tabular data. In Chen et al. \cite{chen_learning_2018}, four data-gen\-er\-at\-ing processes with continuous Gaussian features are created, where not all features contribute to the prediction. They then compare the median rank of informative features regarding different feature attribution methods. Similarly, in Agarwal et al. \cite{agarwal_openxai_2022}, known methods are compared on synthetic and real-world data on a rank level with ground-truth values. However, there are also analyses of model-agnostic feature attribution methods that are more similar to our approach. For example, Guidotti et al. \cite{guidotti_evaluating_2021} create random data-generating processes (DGP) from simple transformations or feature distributions and assess the similarity of methods with the analytical gradients. However, we argue that gradients are not suitable ground-truth values for methods targeting an output decomposition. Further, Liu et al. \cite{liu_synthetic_2021} use additive models from simple feature transformations as DGPs but use Shapley values as ground truth instead. Carmichael et al. \cite{carmichael2023how} propose a framework for comparing the explanations of model-agnostic feature attribution methods with the actual effects of additive structured DGPs.


\section{Understanding the Explanation's Distribution}\label{Sec:Understanding}

To adequately compare feature attribution methods, it is essential to understand their underlying principles before applying them in experiments or benchmark studies. A crucial distinction among these methods lies in how they quantify the effects or influence of features, which can significantly impact visual representations, such as heatmaps or bar plots, and feature rankings, potentially leading to misinterpretations. To illustrate the nuances and a statistical perspective of the state-of-the-art techniques, we consider the following data-generating process (DGP) describing a regression problem:
\begin{align}\label{Sec_3:dgp}
Y = X_1 + X_2 + X_3^2 + X_4 + \varepsilon
\end{align}
where $X_1 \sim \mathcal{N}(0,1)$, $X_2 \sim \mathcal{N}(2, 1)$, $X_3 \sim \mathcal{U}(-1, 2)$, $X_4 \sim \operatorname{Bern}(0.4)$, including Gaussian noise $\varepsilon \sim \mathcal{N}(0,1)$. In this setting, we generally expect a feature attribution method to generate normally distributed relevances for $X_1$ and $X_2$. Similarly, for $X_3$, we expect mostly low with progressively fewer larger values, and for $X_4$, a strongly bimodal distribution. In the following, we group the most well-known feature attribution methods according to their similarities and analyze their distributional and individual behavior using a neural network with ReLU activation trained on $n = 2,000$ instances.


\subsection{Prediction-Sensitive Methods (Group 1)}

\begin{figure}[t]
    \centering
    \begin{subfigure}[b]{0.49\textwidth}
         \centering
         \includegraphics[width=\textwidth]{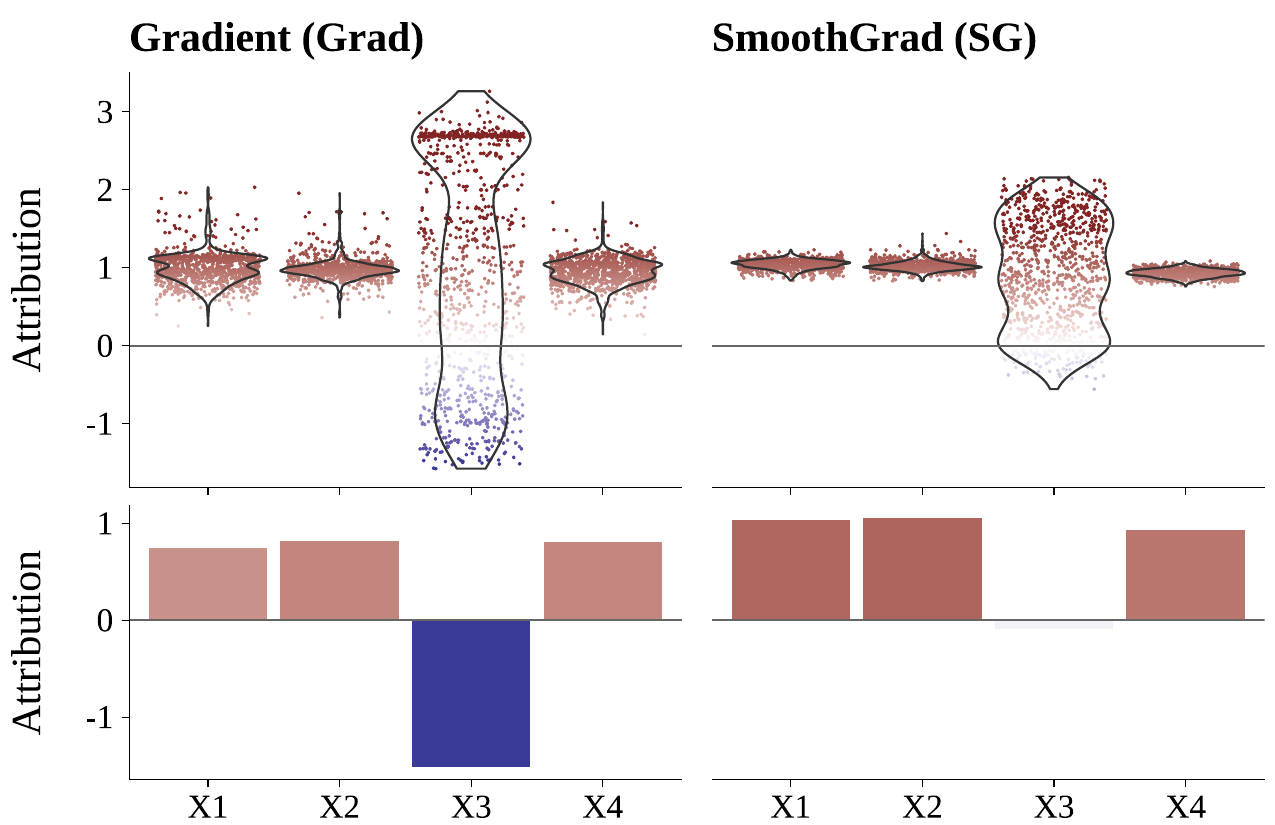}
         \caption{Group 1}
         \label{fig:Sec_3_group_1}
     \end{subfigure}%
     \hfill
     \begin{subfigure}[b]{0.49\textwidth}
         \centering
         \includegraphics[width=\textwidth]{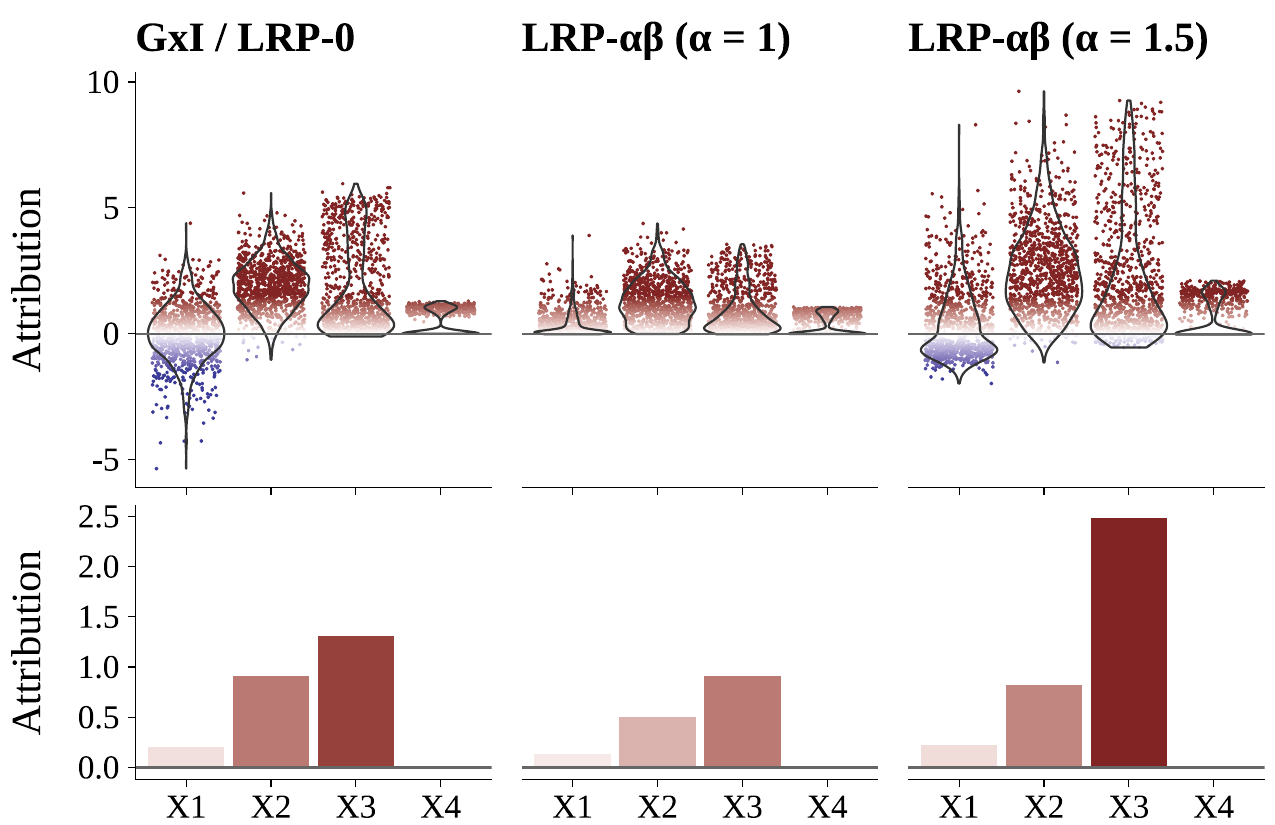}
         \caption{Group 2}
         \label{fig:Sec_3_group_2}
     \end{subfigure}
     \caption{Resulting attribution values of (a) prediction-sensitive and (b) fixed-reference methods of $1,000$ test instances based on the DGP in Eq.~\ref{Sec_3:dgp}. The distribution is shown as a violin plot at the top and a bar plot of the same single instance at the bottom.}
     \label{fig:Sec_3_group_12}
\end{figure}

The first group consists of methods relying on plain gradients, such as the gradient (Grad) \cite{simonyan_deep_2013} method and its variant SmoothGrad (SG) \cite{smilkov_smoothgrad_2017}. Due to their mathematical definition, both methods calculate the output sensitivity of the features, causing them to be unsuitable as local attribution methods for individual effects. For instance, in Figure~\ref{fig:Sec_3_group_1}, it can be seen that both methods in our regression example consistently assign a relevance of closely one for the linear effects of $X_1$, $X_2$, and $X_4$, and almost uniformly distributed relevances for the quadratic effect of $X_3$. Although SmoothGrad visibly reduces the variance of the Grad method, both methods fail to provide appropriate values for the local effects on the prediction, instead indicating the model's sensitivity to changes in the variables. Despite not being further examined in this paper, other plain gradient-based methods like VarGrad \cite{adebayo_local_2018}, FullGrad \cite{srinivas_full-gradient_2019}, and GuidedBackprop \cite{springenberg_striving_2014} fall in this group.

\subsection{Fixed-Reference Methods (Group 2)}

In the second group, we encompass methods that strive to approximate a decomposition of the prediction $f(\bm{x})$ into feature-wise effects. These techniques mainly rely on a first-order Taylor approximation, where the reference point $\bm{\tilde{x}}$ is implicitly fixed. Consequently, the quality of the approximation can significantly depend on the proximity between $\bm{x}$ and $\bm{\tilde{x}}$. The most straightforward variant, Gradient$\times$Input (GxI) \cite{shrikumar_not_2016}, directly extends the Grad method by calculating the Hadamard product of the gradient and the corresponding input features. In a similar way as SG extends the Grad method, GxI can also be extended to SmoothGrad$\times$Input (SGxI), which inherits its smoothing effects. In contrast, the backpropagation-based method layer-wise relevance propagation (LRP) \cite{bach_pixel-wise_2015} represents a sequence of first-order Taylor approximations applied in each individual layer \cite{montavon_explaining_2017}. This involves a layer-wise redistribution of relevances from the upper to lower layers of the model until the input features are reached. Various rules have been proposed for this relevance redistribution, which can also be selected based on the layer type \cite{kohlbrenner2020towards}. In addition to the initial LRP-$0$ rule, there is the LRP-$\varepsilon$ rule, which incorporates a relevance-absorbing stabilizer $\varepsilon > 0$, and the LRP-$\alpha\beta$ rule, which assigns different weights to positive and negative relevances using $\alpha, \beta \in \mathbb{R}$ with $1 = \alpha + \beta$. Setting the $\alpha$ parameter to one (i.e., retaining only positive relevances) and considering solely positive input values makes LRP-$\alpha\beta$ equivalent to the deep Taylor decomposition (DTD) method \cite{montavon_explaining_2017}. For a mathematical description and a more comprehensive overview, we refer to the work by Montavon et al. \cite{montavon_layer-wise_2019}. Despite their differing calculation approaches, GxI and LRP-$0$ implicitly use a reference value of $\bm{0}$ for the Taylor approximations. Moreover, Ancona et al. \cite{ancona_towards_2018} demonstrated that LRP-$0$ and GxI produce identical results for ReLU networks. Furthermore, the LRP-$\alpha\beta$ method and DTD can be interpreted as employing a root point as the reference value, i.e., $f(\tilde{x}) = 0$ \cite{montavon_explaining_2017}. Returning to our regression problem, we observe that GxI precisely yields the expected distribution of effects (see Fig.~\ref{fig:Sec_3_group_2}) regarding a zero baseline. Since we exclusively utilized ReLU activations for the model, this method is equivalent to LRP-$0$. Conversely, with the LRP-$\alpha\beta$ method, one can observe varying weights' influence on positive and negative relevances. When propagating purely positive relevances, all negative relevance values represented in the GxI are truncated to zero. On the other hand, one can see the stretching of positive relevance and compression of negative relevance with a positive-favored propagation in LRP-$\alpha\beta$. At this point, it becomes apparent how different the whole explanation's distribution and the visualizations for individual instances can be, as exemplified by features $X_1$ and $X_3$ in Figure~\ref{fig:Sec_3_group_1} compared to Figure~\ref{fig:Sec_3_group_2}.


\subsection{Reference-Based Methods (Group 3)}

\begin{figure}[t]
    \centering
    \begin{subfigure}[b]{0.49\textwidth}
         \centering
         \includegraphics[width=\textwidth]{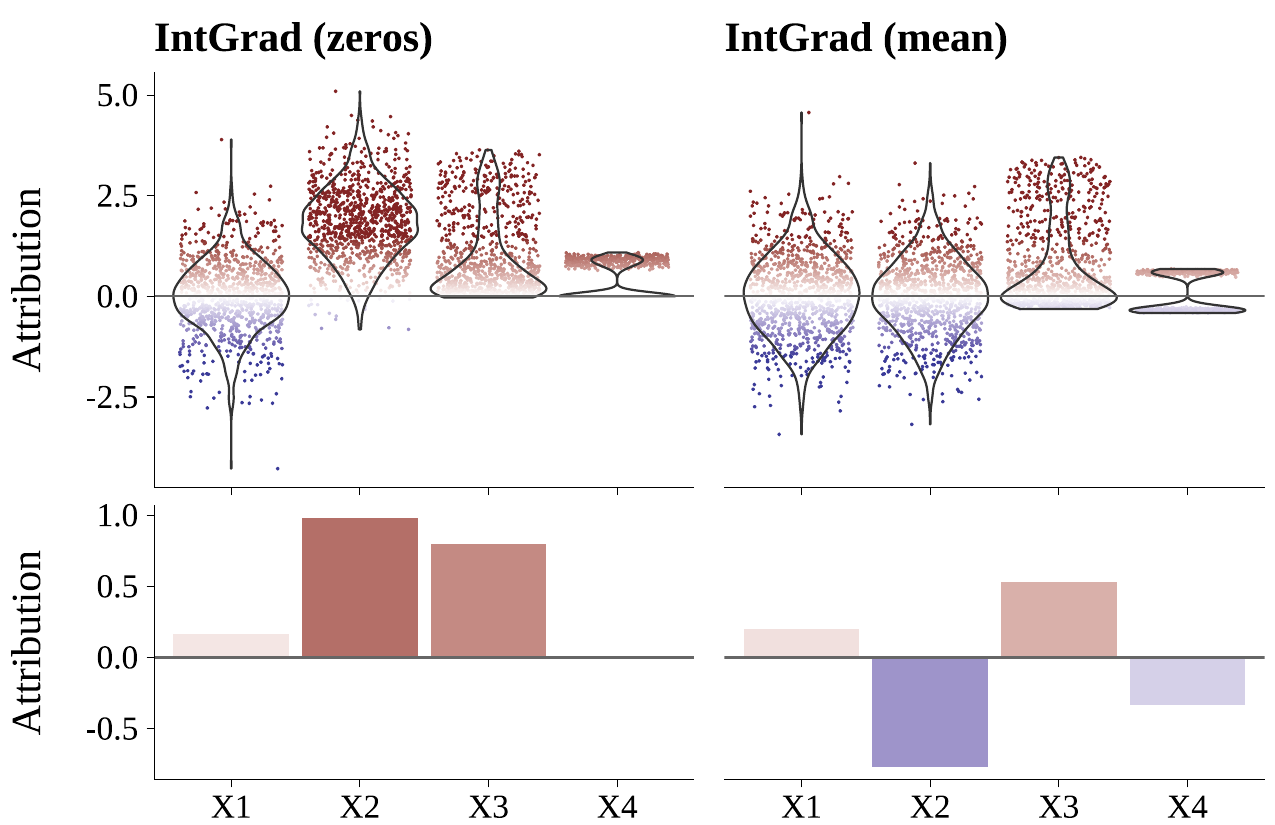}
         \caption{Group 3}
         \label{fig:Sec_3_group_3}
     \end{subfigure}%
     \hfill
     \begin{subfigure}[b]{0.49\textwidth}
         \centering
         \includegraphics[width=\textwidth]{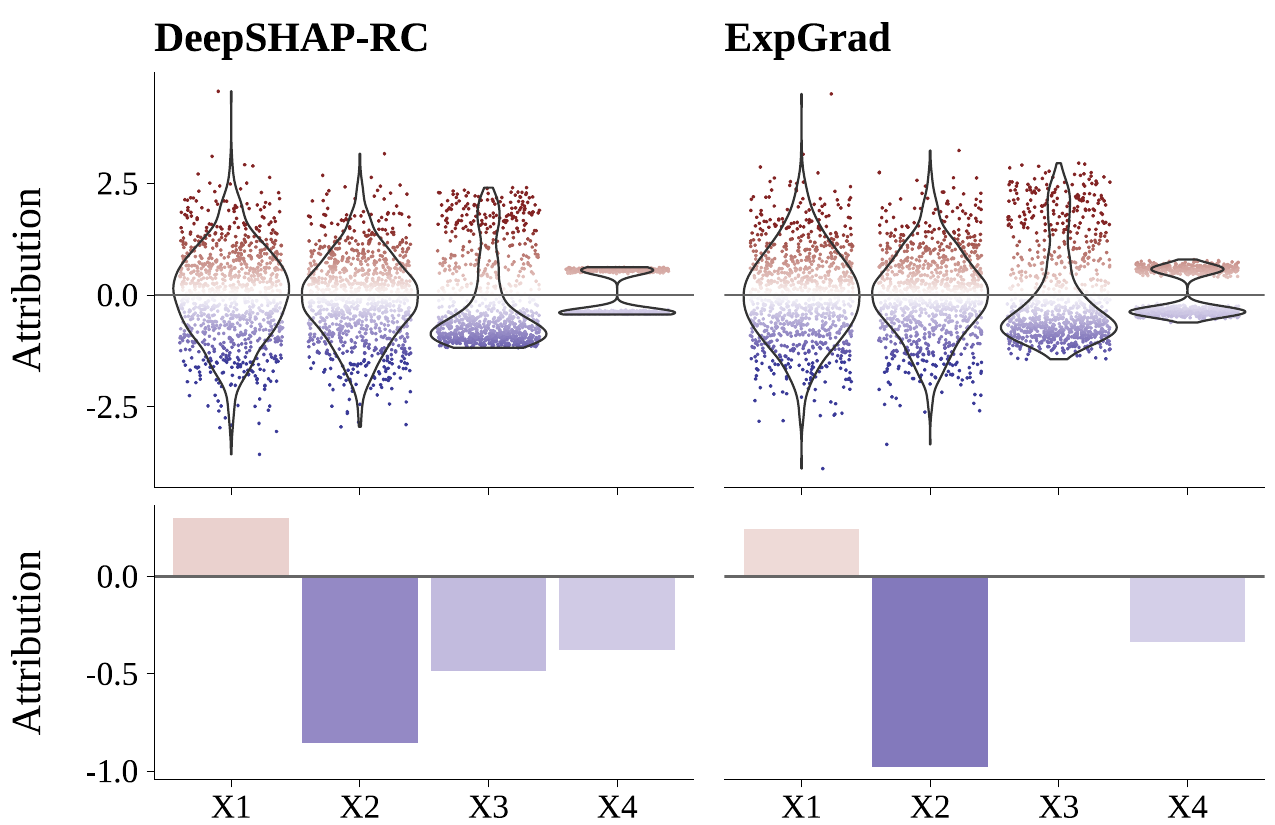}
         \caption{Group 4}
         \label{fig:Sec_3_group_4}
     \end{subfigure}
     \caption{Resulting attribution values of (a) reference-based and (b) Shapley-based methods of $1,000$ test instances based on the DGP in Eq.~\ref{Sec_3:dgp}. The distribution is shown as a violin plot at the top and a bar plot of the same single instance at the bottom.}
     \label{fig:Sec_3_group34}
\end{figure}

The main distinction between the previously discussed methods and our third group is how the local effects are measured. Whereas the former assesses effects with respect to zero or another implicitly defined baseline, the latter group attributes the relative effect of features $\bm{x}$ in relation to an arbitrarily chosen reference value $\bm{\tilde{x}}$. For instance, feature $X_2$ theoretically returns an effect of $2$ on average in the GxI method assuming a perfect model fit, which is a valid explanation based on our data-generating process in Eq.~\ref{Sec_3:dgp}. However, the third group attributes the relative effect of $\bm{x}$ with respect to the baseline value $\bm{\tilde{x}}$, e.g., the average feature value. They aim to decompose the output differences $f(\bm{x}) - f(\bm{\tilde{x}})$ and consequently answer the question of how significant the feature's contribution is compared to the contribution of the reference value. The most prominent methods following this underlying characteristic are integrated gradient (IntGrad) \cite{sundararajan_axiomatic_2017} and deep learning important features (DeepLIFT) \cite{shrikumar_learning_2017}. The former integrates the gradients along a path from $\bm{x}$ to the reference value $\bm{\tilde{x}}$. While this integral is discretized and approximated in applications, an exact decomposition of the difference is asymptotically guaranteed for models that are differentiable almost everywhere. The DeepLIFT method achieves this decompositional target by incorporating the respective layer's intermediate reference value in the layer-wise backpropagation scheme, akin to LRP-$0$. Additionally, the authors propose two rules for propagating through activation functions: the rescale (-RE) and reveal-cancel (-RC) rules. Ancona et al. \cite{ancona_towards_2018} also demonstrated that DeepLIFT-RE using a zero-baseline is equivalent to the computationally efficient GxI method in neural networks using only ReLU activations and with zero bias vectors. Furthermore, empirical evidence from the authors and other researchers indicates a remarkably similar behavior of IntGrad and DeepLIFT-RE in practice \cite{shrikumar_learning_2017,ancona_towards_2018}. In our regression example, we similarly observe this phenomenon and, hence, present only the results of the IntGrad method for the reference value of zeros (left) and of the feature-wise empirical means (right) in Figure~\ref{fig:Sec_3_group_3}. With a zero baseline, the similarity to the GxI method in Figure~\ref{fig:Sec_3_group_2} is clearly visible. However, when the reference value is set to the mean feature values, changes in features $X_2$, $X_3$, and $X_4$ become apparent due to the negative shift. Particularly evident in the second variable is the influence of the reference value, addressing different questions in our regression setting. With $\bm{\tilde{x}} = 0$, the contribution of the variable to the prediction is determined, whereas with $\bm{\tilde{x}}$ set to the average feature value, the effect is attributed relative to the effect of $\bm{\tilde{x}}$ (see Fig.~\ref{fig:Sec_3_group34} bottom variable $X_2$). This shows, in particular, that features with a high assigned magnitude can generally be less relevant or even irrelevant for another baseline.


\subsection{Shapley-based Methods (Group 4)}

The final group consists of methods based on the game-theoretic Shapley values \cite{shapley1953value} adapted for feature attribution in machine learning. They aim to quantify the contribution of each feature to the change in the model prediction concerning the average prediction. Shapley values are computationally expensive to calculate due to the consideration of all possible feature combinations, particularly with high-dimensional data. Therefore, DeepSHAP and expected gradient (ExpGrad) \cite{lundberg_unified_2017,chen2021explaining}, also known as GradSHAP, have been developed for neural networks to approximate these values efficiently. These methods build upon previously explained techniques like DeepLIFT and IntGrad, providing a feature-wise decomposition of predictions compared to the average model prediction. Thus, they represent the feature effect compared to the estimated marginal effect of the feature, i.e., they incorporate the whole feature distribution. These methods are computed by averaging the DeepLIFT or IntGrad results across various reference values. The reference values should originate from the same distribution as the training data, thereby addressing the out-of-distribution problem associated with dataset-independent choices in the plain methods. For example, the mean feature values don't necessarily follow the data distribution. However, this approach entails higher computational costs due to the evaluation of the representative sample of baseline values. Especially for ExpGrad, which involves both aggregating over samples and approximating integrals, Erion et al. \cite{erion2021improving} proposed a purely sample-based estimation approach. In Figure~\ref{fig:Sec_3_group_4} of our running example, only minor differences between the two methods are apparent. However, compared to all other methods, it is clearly evident that the explanation distributions are centered around zero, thus consistently attributing the effect relative to the marginal effect.


\section{Do Feature Attribution Methods Attribute?}

In the preceding section, we demonstrated the varying behavior of the state-of-the-art feature attribution methods, particularly in scenarios where quadratic effects are present in the data-generating process (DGP) or where the feature distribution is not mean-centered. However, we observed that, at least in simple cases, the explanations' distributions are often proportional. This implies that, while the mean and scale of the distribution may differ due to different baselines, the relative distances of explanations should remain consistent across methods. To assess this fundamental behavior of the methods for different types and strengths of effects, we employ an additive data-generating process in the following simulations:
\begin{align}\label{Eq:DGP_simulations}
    Y = \beta_0 + g(X_1)\, \beta_1 + \ldots + g(X_p)\, \beta_p  + \varepsilon.
\end{align}
In this DGP, the function $g: \mathbb{R} \to \mathbb{R}$ determines the type of effect (e.g., $g(x) = x$ for linear or $g(x) = x^2$ for quadratic effects), and the coefficients $\beta_1, \ldots, \beta_p \in \mathbb{R}$ control the feature-individual (global) strength of the effect. Additionally, we add a standard normal distributed error term $\varepsilon \sim \mathcal{N}(0,1)$. Numerical data is sampled from a normal distribution with uniformly sampled mean and variance and then transformed with a linear, piece-wise linear, and non-continuous function $g$. We use equidistant effects from $-1$ to $1$ for the levels of categorical variables. After the data generation, neural networks are trained on this data, and the corresponding feature attribution methods are applied to the test data. See Appendix~\ref{App:sim_details} for more simulation details. The efficacy of a method is evaluated by computing the Pearson correlation between the feature-wise effects $g(x_j^{(1)})\beta_j, \ldots, g(x_j^{(n)})\beta_j $ and the corresponding generated explanations across the test dataset of $1,000$ samples. Consequently, we measure the explanation's distributional fidelity to the shape of the ground-truth effects, still allowing potential linear transformations within the distributions.


\subsection{Impact of Data Preprocessing}
\label{Sec:Preprocessing}

For image data, it is already known that not all feature attribution methods are invariant to constant shifts in the input data \cite{kindermans_reliability_2019}. In particular, for reference-based methods, the question of a suitable baseline has arisen in recent years \cite{haug_baselines_2021}. However, how the data is preprocessed is closely linked to this question. As seen in Section~\ref{Sec:Understanding}, for mean-centered variables, many methods coincide, or shifted variables can be transformed back with the appropriate baselines. Furthermore, to our knowledge, no one has yet analyzed the influence of different encoding techniques for categorical variables on the quality of explanation.

\begin{figure}[!b]
    \centering
    \includegraphics[width = \textwidth]{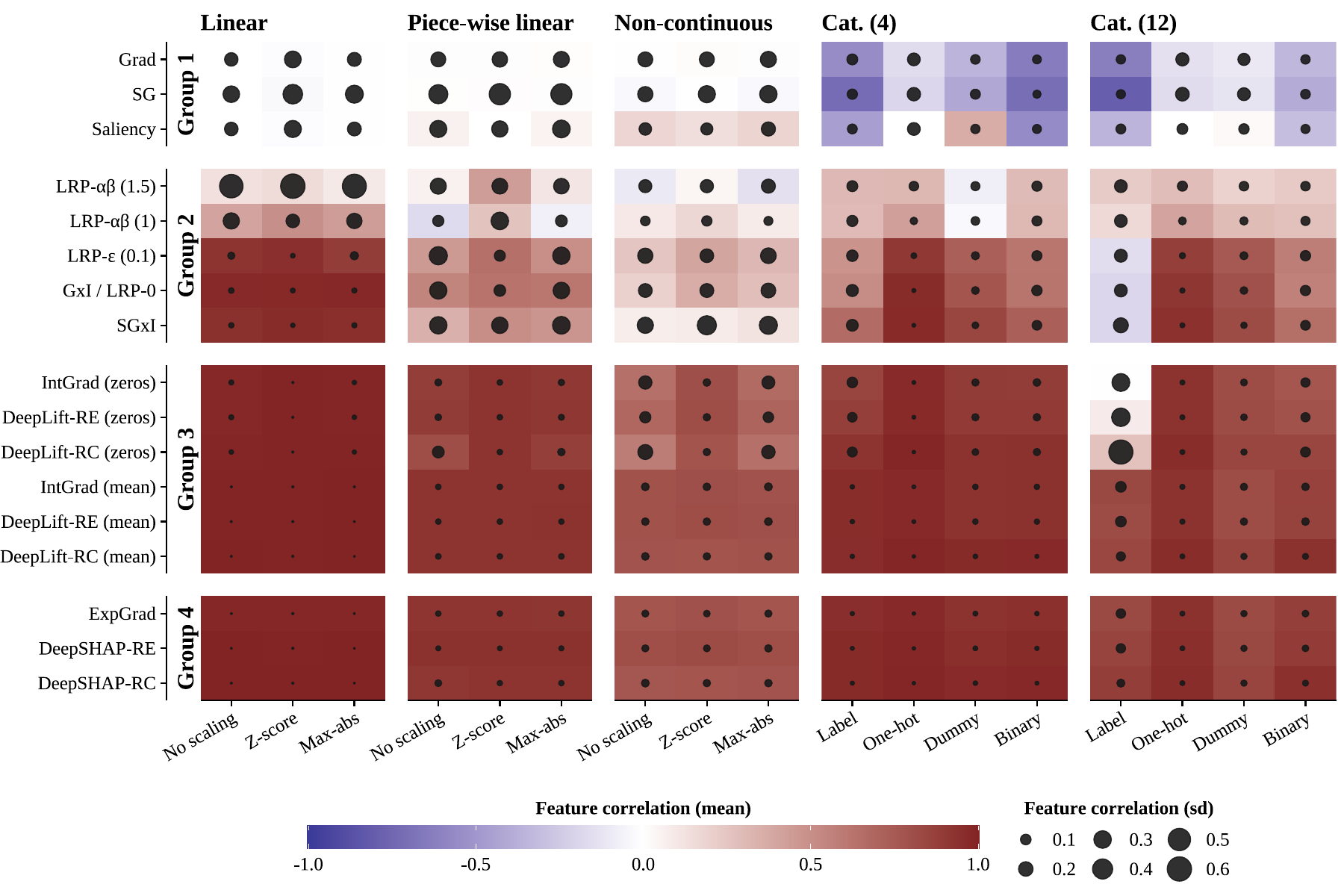}
    \caption{Results of the preprocess simulations for state-of-the-art feature attribution methods (y-axis) showing the averaged correlation with the ground-truth effects across $200$ repetitions and $p=12$ features with equal effect strengths. The individual columns represent different types of effects, and the x-axis shows various preprocessing functions. The size-varying dot describes the standard deviation of the aggregation.}
    \label{fig:Sec_4_Prep_cont}
\end{figure}

In our simulation setup for continuous variables, we consider three common scaling methods: none, z-score, and max-abs scaling. Without scaling, the variables are passed unprocessed to the neural network. In contrast, with the z-score method, the empirical mean is subtracted and then divided by the empirical standard deviation for each variable, resulting in zero-centered and unit-variance data. The min-max preprocessing is particularly well-known for image data, as pixel values are naturally bounded. Nevertheless, this scaling technique is also used for other continuous input data, restricting the data to values between $-1$ and $1$ by scaling with the feature-wise maximum absolute values. In particular, all these scaling methods are invariant to the correlation metric as they describe linear transformations. The scaling parameters are calculated based on the training data and then also used for the test data. For categorical variables, we employ the following encoding techniques: label, one-hot, dummy, and binary encoding. With label encoding, the levels are naively mapped to integer numbers, inducing a particular non-existent order of values. One-hot encoding creates a zero-or-one column for each of the $C$ categories, while dummy encoding transforms a category into $C-1$ columns to avoid redundancy. On the other hand, binary encoding represents each category as a binary number, and then each digit is transformed into a column. This results in a relatively low number of columns, especially for a high number of levels, i.e., only $\lfloor \frac{\log(C)}{\log(2)} + 1 \rfloor$ instead of $C$. For our DGP from Equation~\ref{Eq:DGP_simulations}, we use $p = 12$ variables with identical effect sizes for the two cases of all continuous and all categorical variables, allowing correlations to be unbiasedly aggregated across all variables. For the categorical variables, we consider both a small number of levels ($4$) and a high number of levels ($12$). Each setting is repeated 200 times, and then the mean correlation across the repetitions and features is calculated and summarized in Figure~\ref{fig:Sec_4_Prep_cont}. In addition, we calculate the standard deviation, which is shown next to the aggregated correlation as a size-varying dot.

Generally, the results from Figure~\ref{fig:Sec_4_Prep_cont} need to be interpreted column-wise since each column describes either a different DGP or a different model performance caused by another preprocessing function. However, the models performed notably similarly for continuous variables in the individual effect groups and within a small error range for the categorical variables (see Table.~\ref{tab:model_fit_cont} and \ref{tab:model_fit_cat} in the Appendix). As expected, the prediction-sensitive methods (Grad, SG, and Saliency) are unable to correctly assign proportional local effects for any of the considered effect types or even result in misinterpretations due to negative correlations for categorical variables (see rows of Group 1 methods in Fig.~\ref{fig:Sec_4_Prep_cont}). In addition, the larger dot describing the standard deviation across the simulation's repetitions and features shows an inconsistent attribution and, thus, a strong model dependency. Similarly, the uneven weighting of positive and negative relevances in LRP-$\alpha\beta$ proves to be counterproductive or even resembles guessing for non-linear effects or dummy-encoded variables. The remaining Group 2 methods (LRP-$0$/GxI, LRP-$\varepsilon$, and SGxI) show only slight deviations from Group 3 and 4 methods concerning linear effects and categorical variables. However, their performance significantly worsens and destabilizes when confronted with non-linear effects. For the zero-baseline methods, no scaling and max-abs scaling increasingly degrade the quality and stability with the complexity of the effect types. This phenomenon is also the case for label-encoded categorical variables. This could be mainly due to the fact that zero is no longer a distribution-neutral baseline value for max-abs scaling or label encoding and thus reduces the quality of the Taylor approximation. However, this can be adjusted by using the feature-wise mean value as reference (see lower rows for Group 3 methods in Fig.~\ref{fig:Sec_4_Prep_cont}). Otherwise, there is hardly any difference between the methods in Group 3 with empirical means as reference values, and Shapley-based methods (Group 4), which consistently show high correlations with the ground truth. Nevertheless, the overall results show that, with the exception of the prediction-sensitive methods and LRP-$\alpha\beta$, the methods mostly agree using the zero-centered scaling and common encoding techniques.


\subsection{Faithfulness of Effects}
\label{Sec:Faithfulness}

\begin{figure}[!t]
    \centering
    \includegraphics[width = \textwidth]{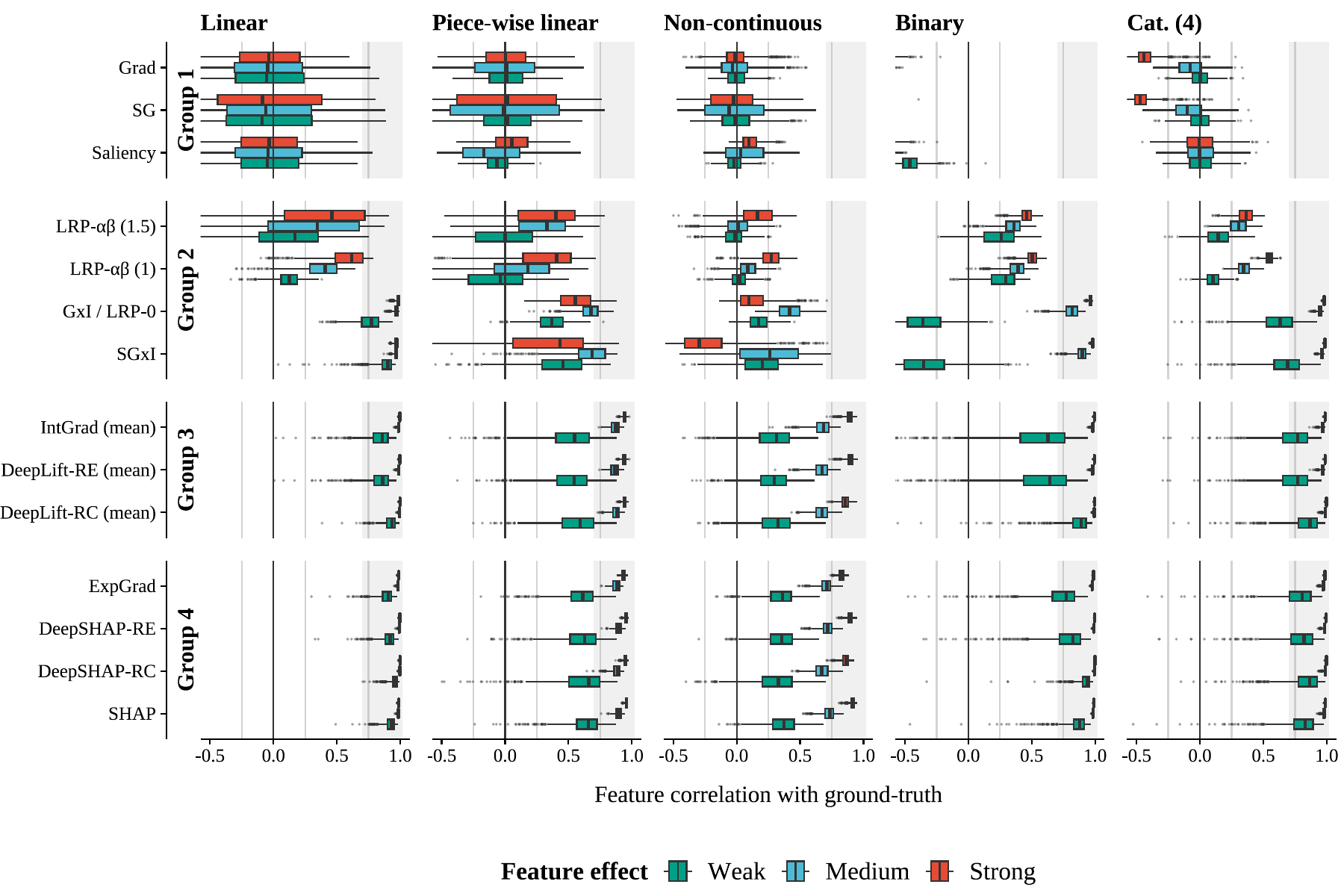}
    \caption{Results of the faithfulness simulations for state-of-the-art feature attribution methods (y-axis) showing the correlation with the ground-truth effects across $200$ repetitions and $p=12$ features with grouped (weak, medium, strong) effect strengths as box plots. The individual columns represent different types of effects, and the x-axis shows the correlation.}
    \label{fig:Sec_Faith}
\end{figure}

Which method to choose is strongly debated in the literature, with disagreement regarding which method provides adequate explanations and which does not. Commonly, the most influential features are removed, and the impact on prediction is evaluated. However, as seen in Section~\ref{Sec:Understanding}, the strength measured as the absolute magnitude of a variable varies and is baseline-dependent. Nevertheless, by adjusting $\beta_1, \ldots, \beta_p$, we want to investigate how the methods behave with different effect sizes. Since weak effects are also more difficult for the model to capture due to a small signal-to-noise ratio, we expect that a good method also detects them less accurately. Similarly to the setting in Section~\ref{Sec:Preprocessing}, we simulate $p = 12$ normally distributed, binary, and categorical variables to evaluate the performance of feature attribution methods concerning different effect sizes. In this scenario of the DGP from Equation~\ref{Eq:DGP_simulations}, grouped variables are considered, where $\beta_1, \ldots, \beta_4 = 0.1$ for weak, $\beta_5, \ldots, \beta_8 = 0.4$ for medium, and $\beta_9, \ldots, \beta_{12} = 1$ for strong effects. Furthermore, we employ z-score scaling for continuous variables, label encoding for binary, and one-hot encoding for categorical variables with four levels. The results of this simulation, conducted over $200$ trained neural networks and $n = 2,000$, are illustrated in Figure.~\ref{fig:Sec_Faith}. For a comparison with a state-of-the-art model-agnostic method, we also execute the well-established SHAP method \cite{lundberg_unified_2017} on the test data, which approximates Shapley values.

The results for the prediction-sensitive methods are similar to those for the preprocessing simulation: The continuous variables tend to have a strongly varying correlation around zero for all effect strengths and types, which represents a guessing of the local contributions to the prediction. The strong negative correlation for binary variables is probably due to unusual inter- or extrapolations of the model around the values $0$ and $1$. For the Group 2 methods, LRP-$\alpha\beta$ moderately captures the ground-truth attributions for linear data but increasingly fails to provide reliable explanations for more complex relationships. The method GxI (and thus LRP-$0$) appears to provide relevances quite similar to ground truth for (partially) linear effects. However, it only yields moderate correlations for non-linear effects and struggles to recognize the association with strong effect sizes. This inconsistency likely stems from the method's reliance on the first-order Taylor approximation, causing it to falter in non-linear relationships. Interestingly, the GxI and SGxI methods inherit the negative correlation from the gradients for binary variables with weak effects, but this is corrected for one-hot encoded variables. This could presumably be because the Taylor approximation collapses when zero is used as both the reference and explained value. 
As for reference-based and Shapley-based methods, it becomes evident that these methods accurately identify the actual effects across all considered settings. However, Shapley-based methods tend to attribute weak or moderate effects better than all other methods. Furthermore, DeepLIFT-RC and its extension DeepSHAP-RC demonstrate an outstandingly strong correlation with the ground-truth effects, especially for binary and categorical variables. The comparison with the Shapley values from the model-agnostic method SHAP further indicates that the Group 3 and 4 methods align with this popular XAI method and provide a fast and accurate approximation of Shapley values for neural networks.


\subsection{Beyond Feature Attribution Toward Importance}
\label{Sec:Beyond_attribution}

The previous simulations have shown to what extent feature attribution methods are able to attribute the exact proportions of a variable's contribution to the prediction, i.e., how adequate the explanation's distribution aligns with the ground-truth effects. As shown in Section~\ref{Sec:Understanding}, it can now occur that a feature reflecting the mean value correctly, receives a relevance score close to zero if the effect is measured to a zero baseline, even though it is a very important feature from a global perspective. This is comparable to a Gaussian-distributed variable in a linear model with a high regression coefficient: although the global importance of this variable is very high, the attributed relevance for feature values close to zero can vanish. From a local perspective, this means that instance-wise feature rankings based on the magnitude of relevance are not suitable for answering whether one feature is more important than another. Instead, the feature attribution methods do what they are developed for and give the local effect to the prediction relative to a method-dependent baseline. However, all methods should follow the property of \emph{consistency} and assign little to no relevance to unimportant features, while tending to assign higher relevances to crucial features across the dataset. Thus, ranked relevances are more appropriate to answer whether or not a feature is important.

To evaluate the effectiveness of the methods in identifying important and unimportant features, we simulate $p = 20$ normally distributed and categorical variables with $4$ levels using the DGP from Equation~\ref{Sec_3:dgp}, where only the even feature indices affect the regression outcome. For each instance from the test data, the ranking of absolute explanations is converted into the binary decision of whether the feature belongs to the top $10$ most important ones. Since the ground-truth importance values are $1$ for even and $0$ for odd indices, we compute the F1-score for each instance and then average the scores over the $1,000$ test instances. Additionally, we vary the sample size $n$ of the training data to assess the question of detecting important/unimportant features at different model qualities.

\begin{figure}[t]
    \centering
    \includegraphics[width = \textwidth]{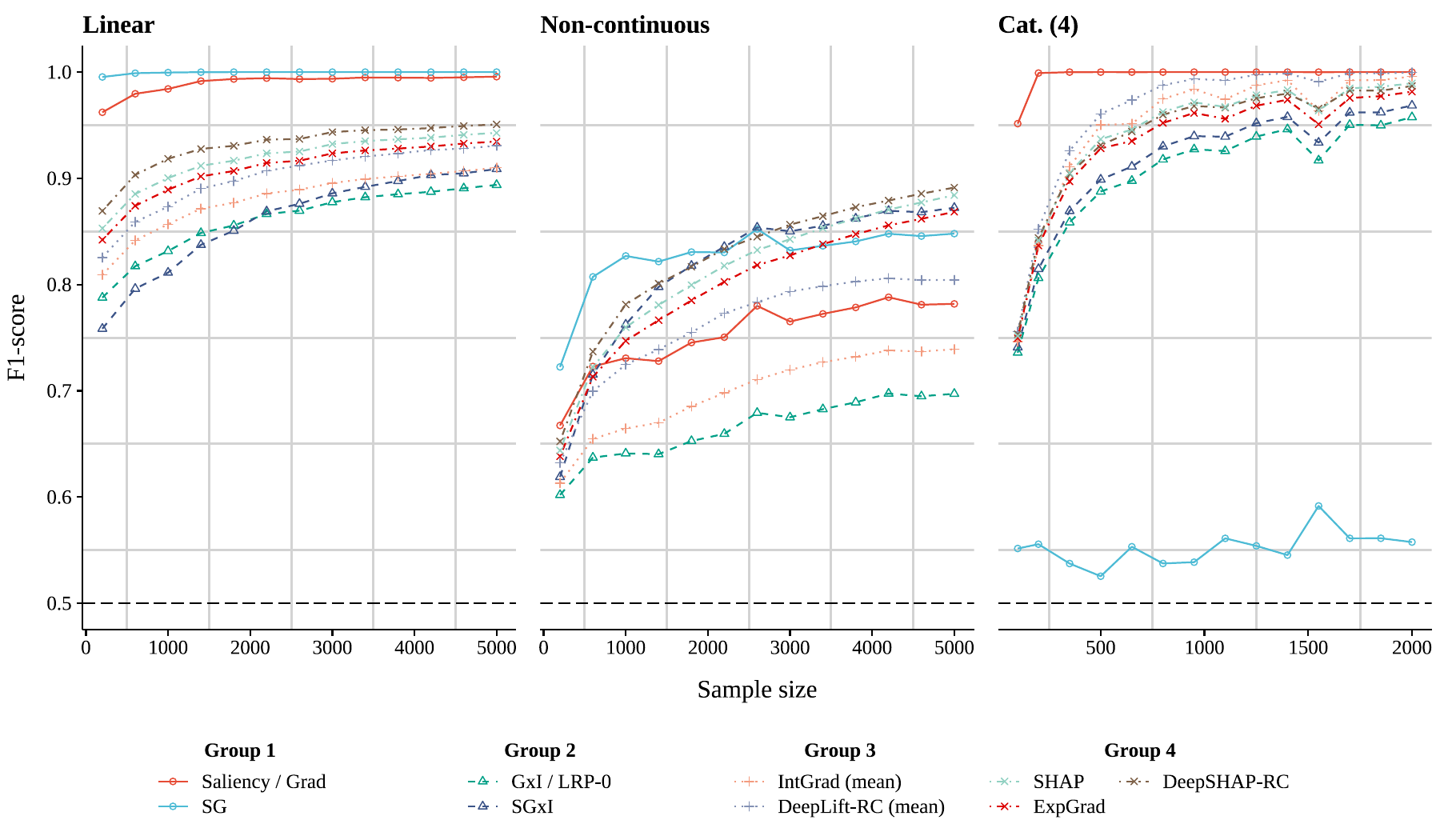}
    \caption{The figure shows the F1-score on the y-axis depending on the sample size for various feature attribution methods (colors) and effect types (columns) averaged over $500$ repetitions.}
    \label{fig:Beyond_attr}
\end{figure}

Although this group of methods performed poorly in the previous simulations, Figure~\ref{fig:Beyond_attr} demonstrates the strength of prediction-sensitive methods. They are capable of nearly perfect discrimination between important and unimportant features across all model qualities for linear effects and categorical variables. Except for SG for categorical variables, which probably smooths the gradients too much due to the high variance in perturbations. Even with limited training data, Grad and especially SG perform exceptionally well for non-linear effects. Aside from Group 1 methods, it is apparent that Shapley-based methods, particularly DeepSHAP-RC, are the most reliable in distinguishing important and unimportant features. They even outperform Grad and SG when dealing with non-linear effects and large sample sizes.
Despite the Group 3 methods with the feature-mean reference performing nearly identically to the Group 4 methods in the previous simulations (see Sec.~\ref{Sec:Faithfulness}), the single-reference methods show significant weaknesses in discriminating important and unimportant features. This is either due to disrupted rankings resulting from shifted relevances (as observed in Section~\ref{Sec:Understanding}) or because Shapley-based methods attribute low relevances to unimportant features more accurately due to the incorporation of multiple baselines. Furthermore, it is again evident that the methods GxI and LRP-$0$ encounter significant issues with non-linear effects and generally perform worse. Moreover, the Shapley-based model-specific feature attribution methods are able to outperform the established model-agnostic method SHAP.

\section{Discussion}

We have demonstrated through simulations how differently state-of-the-art feature attribution methods measure local effects on a prediction and how visualizations could be manipulated in favor of chosen features using baseline values (see Section~\ref{Sec:Understanding}), which can distort the quality of rank-based measures for determining important features. This confirms previous research that has already demonstrated this influence on heatmaps for reference-based methods in applications \cite{sturmfels2020visualizing,mamalakis_2023,dolci_2023}. Closely connected to the choice of baseline values, we -- to the best of our knowledge -- for the first time analyzed the impact of common data preprocessing steps on the quality of attribution for continuous and categorical variables. We found that, except for prediction-sensitive methods, standard normalization techniques such as z-score scaling for continuous and one-hot encoding for categorical variables provide the most accurate attributions of local effects. However, fixed-reference methods such as LRP-$0$, GxI, and LRP-$\alpha\beta$ struggle noticeably with non-linear effects. Nevertheless, our simulations have shown that an appropriate baseline can adjust for non-zero-centered scaling techniques in reference-based methods.

Furthermore, in Section~\ref{Sec:Faithfulness}, we observed that the distributions of explanations provided by reference-based and Shapley-based methods exhibit remarkable similarities for standard preprocessing techniques. These explanations correlate strongly with ground-truth effects as the effect size increases, with Shapley-based methods generally tending to be more accurate. These simulations demonstrate that reference-based, Shapley-based, and fixed-reference-based methods (except for linear transformations) yield proportional explanations' distributions, contradicting the disagreement problem \cite{krishna_disagreement_2022,neely2021order}. Hence, the disagreement is mainly caused by different baselines. However, our findings underscore that feature attribution methods, notably influenced by the choice of baseline, pursue distinct objectives in attribution, consequently resulting in varying local magnitudes of relevances. This fact became particularly evident in Section~\ref{Sec:Beyond_attribution}, where the methods exhibited noticeable differences in discriminating between important and unimportant features based on ranked magnitudes of the explanations. While prediction-sensitive methods may falter in correctly attributing relevances, they performed well in determining whether or not a feature is important. Nevertheless, Shapley-based methods, especially DeepLIFT-RC, appear to consistently excel in addressing this binary classification problem, as also observed by other researchers \cite{zhou_feature_2022}.

While our simulations are based on simple synthetic data without correlated features and interaction effects and we only trained dense neural networks, they represent the first independent comparison of feature attribution methods on tabular data considering the attributions' correlation. Additionally, dense layers serve as a fundamental building block for many modern deep neural networks, such as convolutional neural networks or attention modules, providing insights into their behavior. Furthermore, the restriction to regression problems is negligible, as most feature attribution methods ignore the activation of the final layer, which computes class probabilities, and instead apply the method to the preactivation values \cite{bach_pixel-wise_2015,simonyan_deep_2013,shrikumar_learning_2017}. Investigating the methods for interaction effects remains an attractive direction for future work, building upon the theoretical groundwork already explored by Deng et al. \cite{deng_2024}.

\section{Conclusion}

Our study provides a fundamental understanding of state-of-the-art feature attribution methods for neural networks through simulation studies. Initially, we demonstrated the variability in the explanation's distribution and individual explanations across different methods. Particularly, we highlighted how crucial the implicitly or explicitly set reference value can influence the magnitude of a feature's relevance and, thus, the ranking regarding the importance of a feature on a local level. Furthermore, we illustrated how preprocessing techniques of training data can affect and destabilize the quality of attribution and can only be corrected in reference-based or Shapley-based methods through appropriate baseline values.

Nevertheless, we have shown that most state-of-the-art methods, when utilizing z-score scaling for continuous variables and one-hot encoding for categorical variables, deliver relevances that closely correlate with the ground-truth values as the effect strength increases. However, this comparison does not consider linear transformations of the distributions, leading to the methods' disagreement when transitioning to rank-aggregated values. Additionally, we have demonstrated that plain gradient methods, such as the gradient (Grad) and SmoothGrad (SG), are not suitable as attribution methods for effect decompositions while being highly capable of distinguishing important and unimportant features on a global scale.

\subsubsection*{Acknowledgement}
This project was funded by the German Research Foundation (DFG), Emmy Noether Grant 437611051.

\bibliographystyle{unsrt}  
\bibliography{references}  

\newpage
\appendix

\section{Appendix}

All figures and simulation results presented in this work are reproducible using the code hosted on our GitHub repository, available at \url{https://github.com/bips-hb/Toward_Understanding_Disagreement_Problem}.

\subsection{COMPAS Dataset}
\label{App:COMPAS}

For this example, we load the COMPAS dataset from the R package \texttt{mlr3fairness}\footnote{\url{https://mlr3fairness.mlr-org.com/reference/compas.html}} and train it with respect to the variable \texttt{two\_year\_recid}. We use a neural network model with four layers: $256$, $128$, and $64$ neurons in the hidden layers, along with ReLU activations. Additionally, a dropout layer is added after each hidden layer for regularization. Continuous variables are preprocessed using z-scores, and categorical variables are one-hot encoded. Using an $80/20$ train-test split, we achieve an F1-score of $74.36\%$.

\subsection{Simulation Details}
\label{App:sim_details}

\subsubsection{Data Generation}

An additive model with independent variables is used for all simulations, following the data-generating process from Equation~\ref{Eq:DGP_simulations}. The continuous variable $X_i$ is sampled from a normal distribution $\mathcal{N}(\mu_i, \sigma_i)$, where $\mu_i$ and $\sigma_i$ are uniformly distributed within the range $[-2, 2]$ for the mean and $[0.9, 1.1]$ for the variance. This approach allows us to simulate variations in the mean and scale of the Gaussian distributions. For a categorical variable $X_i$ with $c \in \mathbb{N}$ levels $A_1, \ldots, A_c$ with equal level probabilities, equidistant effects ranging from $-1$ to $1$ are assigned, i.e., $g(X_i = A_k) = -1 + \frac{2(k-1)}{c - 1}$. This zero-centered distribution of effects across categories ensures that only the coefficient $\beta_i$ controls the effect strength so that the level-specific effects do not disturb it.

\begin{wrapfigure}{r}{0.55\textwidth}
  \centering
    \includegraphics[width = 0.50\textwidth]{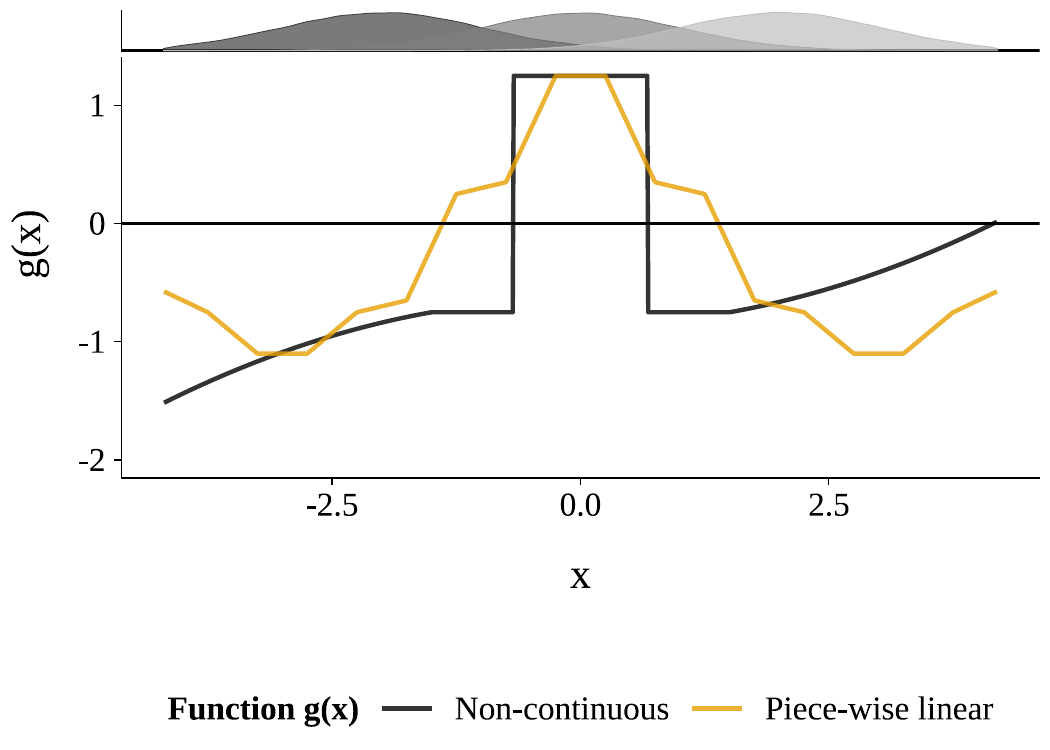}
    \caption{Transformations $g$ used for non-linear relationships of the variables and the regression outcome.}
    \label{fig:app_trans}
\end{wrapfigure}
In order to simulate different types of effects for continuous variables also covering the mean shifts, we consider the linear function $g(x) = x$ and the transformations $g:\mathbb{R} \to \mathbb{R}$ described in Figure~\ref{fig:app_trans}. Unless otherwise stated, $4,000$ training instances are generated for continuous and $2,000$ for categorical/binary variables, with one-third of each generated as evaluation data for the neural network training. The smaller number of samples for the categorical and binary variables is due to their simpler and discrete relationships with the regression outcome. Additionally, we use $1,000$ instances as test data for the feature attribution methods.

\subsubsection{Neural Network Training}

The neural networks consist of three dense layers with $256$, $128$, and $64$ neurons for continuous variables and, due to simpler relationships, $128$, $64$, and $32$ neurons for categorical variables. ReLU is always used as the activation function, and a dropout layer with a dropout rate of $0.4$ is added after the activation in the hidden layers. Each network is trained for a maximum of $300$ epochs using the Adam optimizer, where the initial learning rate of $0.01$ is multiplied by $0.2$ every $50$ epochs, and the training terminates after $50$ unimproved epochs on the evaluation data.

\subsubsection{Hyperparameters of Feature Attribution Methods}

We apply the feature attribution methods using the R package \texttt{innsight} \cite{koenen2023interpreting} on the generated test data and the trained model. No hyperparameters are needed for the methods, Grad, Saliency, and GxI; only the corresponding rules and baselines are required for LRP, DeepLIFT, and DeepSHAP. For SmoothGrad, we take $50$ samples with a noise level of $0.2$. Similarly, we use $50$ samples for IntGrad, ExpGrad, and SHAP methods. Since methods for encoding categorical variables, except for label encoding, generate new artificial variables, we subsequently summed them up, ensuring that each categorical feature is assigned only one relevance.

\subsection{Model Performance}

Since methods for encoding categorical variables, except for label encoding, generate new artificial variables, we subsequently summed them up, ensuring that each categorical feature is assigned only one relevance value (see Table~\ref{tab:model_fit_cont} and \ref{tab:model_fit_cat}).

\begin{table}[!h]
\caption{Results of the neural network performance on test data consisting of continuous variables measured by the average R² value ($\pm$ standard deviation) over $200$ repetitions. As a reference, the performance of a linear model is also included to show that the neural network learned the non-linear relationships.}
\centering
\resizebox{0.95\textwidth}{!}{
\begin{tabular}{*{1}{>{\centering\arraybackslash}p{0.1\textwidth}}*{1}{>{\centering\arraybackslash}p{0.2\textwidth}}*{3}{>{\centering\arraybackslash}p{0.233\textwidth}}}
\toprule
                          &              & \multicolumn{3}{c}{Effect type}             \\ \cmidrule(l){3-5} 
                          & Scaling & Linear & Piece-wise linear & Non-continuous \\ \midrule
\multirow{3}{*}{Sec. 4.1} & No scaling   &  $0.92 \pm 0.01$ &  $0.74 \pm 0.02$ & $0.57 \pm 0.03$  \\
                          & Z-Score      &  $0.91 \pm 0.01$ &  $0.75 \pm 0.02$ & $0.60 \pm 0.03$    \\
                          & Max-Abs      &  $0.92 \pm 0.01$ &  $0.75 \pm 0.02$ & $0.57 \pm 0.04$    \\ 
                          & (Linear model) & $0.92 \pm 0.00$&  $0.47 \pm 0.06$ & $0.23 \pm 0.04$\\ \midrule
Sec. 4.2                  & Z-Score      &  $0.81 \pm 0.01$ &  $0.60 \pm 0.03$ & $0.60 \pm 0.03$  \\ 
                          & (Linear model) & $0.82 \pm 0.01$&  $0.38 \pm 0.08$ & $0.20 \pm 0.06$\\ \bottomrule
\end{tabular}}
\label{tab:model_fit_cont}
\end{table}

\begin{table}[!h]
\caption{Results of the neural network performance on test data consisting of categorical variables measured by the average R² value ($\pm$ standard deviation) over $200$ repetitions.}
\centering
\resizebox{0.95\textwidth}{!}{
\begin{tabular}{*{1}{>{\centering\arraybackslash}p{0.1\textwidth}}*{1}{>{\centering\arraybackslash}p{0.15\textwidth}}*{4}{>{\centering\arraybackslash}p{0.1875\textwidth}}}
\toprule
                          &              & \multicolumn{4}{c}{Encoding}             \\ \cmidrule(l){3-6} 
                          & \# Levels & Label & One-hot & Dummy & Binary  \\ \midrule
\multirow{2}{*}{Sec. 4.1} & 4   &  $0.57 \pm 0.02$ &  $0.60 \pm 0.02$ & $0.55 \pm 0.02$ & $0.56 \pm 0.02$\\
                          & 12  &  $0.37 \pm 0.05$ &  $0.48 \pm 0.03$ & $0.37 \pm 0.03$ & $0.42 \pm 0.03$   \\ \midrule
\multirow{2}{*}{Sec. 4.2} & Binary   &  $0.81 \pm 0.01$ & $-$ & $-$ & $-$ \\ 
                          & 4 & $-$ &  $0.7 \pm 0.02$ & $-$  & $-$\\ \bottomrule
\end{tabular}}
\label{tab:model_fit_cat}
\end{table}

\end{document}